\documentclass[10pt,twocolumn,letterpaper]{article}

\usepackage{wacv}
\usepackage{times}
\usepackage{epsfig}
\usepackage{graphicx}
\usepackage{amsmath}
\usepackage{amssymb}
\usepackage{booktabs}
\usepackage{subfig}
\usepackage[flushleft]{threeparttable}
\usepackage{tablefootnote}
\usepackage{comment}
\def\wacvPaperID{2113} % Enter the WACV Paper ID here

\wacvalgorithmstrack   % Uncomment this line if you are submitting to the Algorithms Track.
\wacvfinalcopy % *** Uncomment this line for the final submission

\ifwacvfinal
\usepackage[breaklinks=true,bookmarks=false]{hyperref}
\else
\usepackage[pagebackref=true,breaklinks=true,colorlinks,bookmarks=false]{hyperref}
\fi

\begin{document}

%%%%%%%%% TITLE
\title{Minutiae-Guided Fingerprint Embeddings via Vision Transformers}

\author{Steven A. Grosz\\
Michigan State University\\
{\tt\small groszste@msu.edu}
\and
Joshua J. Engelsma\thanks{This author's affiliation was with Amazon at the time of writing this paper, but is now with Rank One Computing.}\\
Rank One Computing\\
{\tt\small josh.engelsma@rankone.io}
\and
Rajeev Ranjan\\
Amazon\\
{\tt\small rvranjan@amazon.com}
\and
Naveen Ramakrishnan\\
Amazon\\
{\tt\small navramk@amazon.com}
\and
Manoj Aggarwal\\
Amazon\\
{\tt\small manojagg@amazon.com}
\and
Gerard G. Medioni\\
Amazon\\
{\tt\small medioni@amazon.com}
\and
Anil K. Jain\\
Michigan State University\\
{\tt\small jain@msu.edu}
}

\maketitle
\thispagestyle{empty}

%%%%%%%%% ABSTRACT
\begin{abstract}
   Minutiae matching has long dominated the field of fingerprint recognition. However, deep networks can be used to extract fixed-length embeddings from fingerprints. To date, the few studies that have explored the use of CNN architectures to extract such embeddings have shown extreme promise. Inspired by these early works, we propose the first use of a Vision Transformer (ViT) to learn a discriminative fixed-length fingerprint embedding. We further demonstrate that by guiding the ViT to focus in on local, minutiae related features, we can boost the recognition performance. Finally, we show that by fusing embeddings learned by CNNs and ViTs we can reach near parity with a commercial state-of-the-art (SOTA) matcher. In particular, we obtain a TAR=$94.23\%$ @ FAR=$0.1\%$ on the NIST SD 302 public-domain dataset, compared to a SOTA commercial matcher which obtains TAR=$96.71\%$ @ FAR=$0.1\%$. Additionally, our fixed-length embeddings can be matched orders of magnitude faster than the commercial system (2.5 million matches/second compared to 50K matches/second). We make our code and models publicly available to encourage further research on this topic:~\url{github.com/tba}.
\end{abstract}

%%%%%%%%% BODY TEXT
\section{Introduction}
\begin{figure}[t]
  \centering
  \includegraphics[width=\linewidth]{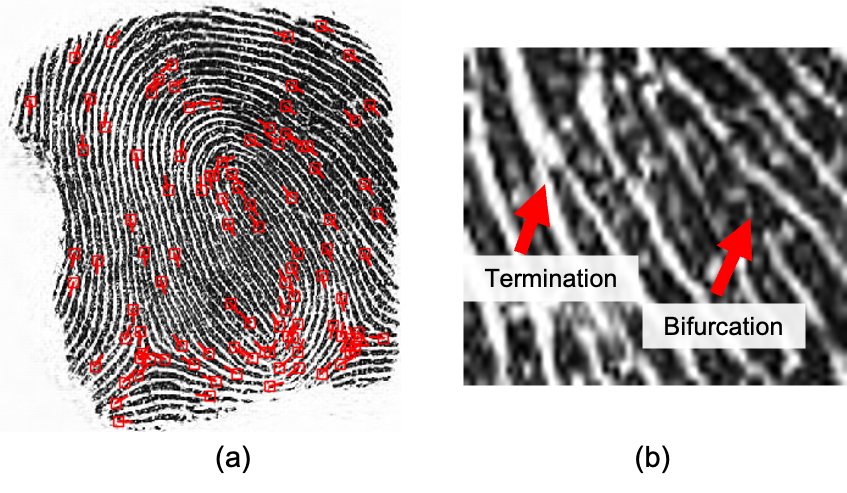}
  % \subfloat[Example PrintsGAN fingerprint annotated with minutiae points]{\includegraphics[width=0.45\linewidth]{figures/mnt_annotation.png}\label{fig:f2}}
  % \hfill
  % \subfloat[Minutiae types]{\includegraphics[width=0.45\linewidth]{figures/minutiae_types.png}\label{fig:f1}}
  \caption{The most prevalent fingerprint representation is comprised of a variable length, unordered minutiae (key-point) set. (a) A full minutiae set from a computer generated (synthetic) fingerprint~\cite{engelsma2022printsgan}. Each minutia point has a location $(x, y)$ and an orientation $\theta$ indicating the position and direction, respectively. (b) Examples of the two types of fingerprint minutiae (Termination and Bifurcation).}
  \label{fig:intro}
\end{figure}

Over the past several decades, fingerprint recognition systems have become pervasive across the globe in a number of different applications, from mobile phone unlock to national ID programs~\cite{maltoni2022handbook}. The widespread adoption of Automated Fingerprint Identification Systems (AFIS) can be primarily attributed to two major tenants: 

\noindent \textbf{Accuracy:} According to the ongoing Proprietary Fingerprint Template III (PFT III~\footnote{\url{https://www.nist.gov/itl/iad/image-group/proprietary-fingerprint-template-pft-iii}}) evaluations conducted by the National Institute of Standards and Technology (NIST), fingerprint recognition systems are now able to obtain recognition accuracies across multiple operational datasets (collected for various use-cases) in excess of 99\%. 
    
\noindent\textbf{Scientific Understanding:} Fingerprints were long believed to be both permanent (retaining the same high accuracy over time) and unique (different for every person, even different fingers of the same person). Rigorous statistical analyses have demonstrated that these central tenants are indeed backed by strong evidence~\cite{yoon2015longitudinal, pankanti2002individuality}.

Both of these tenants have been established primarily via the use of long-standing discriminative features widely known as fingerprint~\textit{minutiae}~\cite{galton}. Minutiae points are anomalous key-points located throughout the fingerprint's friction ridge pattern. These anomalies occur as either i) terminations or ii) bifurcations (see (b) of Figure~\ref{fig:intro} for examples). Furthermore, each minutiae point is a 3-tuple of $(x, y, \theta)$ where $x, y$ indicate the location of the minutiae point and $\theta$ is the direction of the ridge flow at the minutiae point's location. 

\begin{figure}[t]
 \centering
 \includegraphics[scale=0.25]{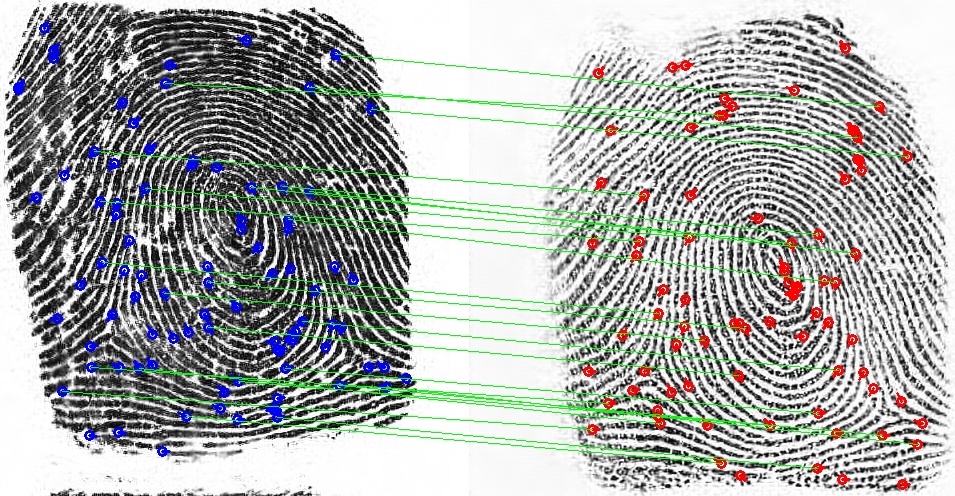}
 \caption{An example illustrating minutiae correspondences between a pair of synthetic fingerprints~\cite{engelsma2022printsgan} of the same finger. A total of 23 minutiae points are in correspondence shown with green lines. Correspondences were automatically established using the graph matching algorithm from~\cite{cao2019end}}
 \label{fig:corr}
 \vspace{-1.5em}
\end{figure}

In almost all state-of-the-art (SOTA) fingerprint recognition systems, a full minutiae set (shown in (a) of Figure~\ref{fig:intro}) is first extracted from a given fingerprint. Subsequently, this variable length, unordered set of minutiae key-points is compared to another set of minutiae key-points extracted from an enrolled fingerprint image using graph matching techniques (Figure~\ref{fig:corr}). At the most basic level, if a similarity score aggregated from corresponding points is more than a specified threshold, the fingerprint pair is determined to be a genuine match, otherwise a non-matching imposter pair. 

Although the success of the minutiae based features have led to \textit{minutiae} and \textit{automated fingerprint recognition} being nearly synonymous terms, minutiae based fingerprint matching systems do have several significant limitations:

\noindent\textbf{Inefficiency:} Minutiae matching is computationally expensive. First minutiae must be detected. Oftentimes, descriptors are also extracted for each minutiae point. Then, these variable length, unordered sets need to be compared via expensive graph matching techniques. In contrast, most SOTA face recognition systems which rely on deep feature representations require only a dot product between fixed-length embeddings of query and enrollment images to compute a match score ($d$ multiplications and $d-1$ additions for a $d$ dimensional embedding).
\noindent\textbf{Vulnerability:} Matching fingerprint templates\footnote{We use the term templates, representations, and embeddings throughout to denote a set of features extracted from a fingerprint image.} in a secure encrypted manner is extremely challenging. In contrast, fully homomorphic encryption (FHE) schemes on deep embeddings have now shown the ability to match biometric templates in the encrypted domain in real-time~\cite{boddeti2018secure, engelsma2022hers}.

\newcommand{\specialcell}[2][c]{%
  \begin{tabular}[#1]{@{}c@{}}#2\end{tabular}}
  \newcommand{\tabitem}{~~\llap{\textbullet}~~}

These limitations of the minutiae template and also the demonstrated success of SOTA deep networks to extract highly discriminative biometric embeddings from faces~\cite{schroff2015facenet, yi2014learning}, has initiated exploration in alternative representations to fingerprint minutiae. In particular, works from~\cite{cao2017fingerprint, song2017fingerprint, song2019aggregating, li2019learning, takahashi2020fingerprint, gu2022latent} all explore the use of deep networks to embed fingerprint images into a compact, discriminatory, fixed-length fingerprint representation. These works show tremendous promise in terms of both the accuracy and speed needed to either supplant or at least complement the established minutiae template. For example, Engelsma \textit{et al.} showed in~\cite{engelsma2019learning} that a $192$-dim deep fingerprint embedding could reach near parity with COTS matchers in terms of authentication and search accuracy on NIST SD4~\cite{sd4} and NIST SD14~\cite{sd14} datasets while matching at 3 or 4 orders of magnitude faster speed (300 milliseconds search time vs. 27,000 milliseconds search time on a gallery of 1.1 million fingerprints). On resource constrained devices and for civil ID and law enforcement databases with hundreds of million images in the datasets, these improvements in search time are invaluable. 

Given the complementary strengths of the minutiae template and the deep fingerprint templates (human interpretability, statistical understanding, and interoperability of the minutiae template) vs. (accuracy and speed of the deep templates), a natural idea is to learn a fingerprint embedding which somehow distills knowledge of fingerprint minutiae into the parameters of the deep network. Rather than completely discarding the minutiae template, we lean on this domain knowledge to learn more discriminative and generalizable deep fingerprint embeddings. Indeed, the models in~\cite{engelsma2019learning, song2019aggregating, takahashi2020fingerprint} all aim to do just that using Convolutional Neural Networks (CNN) in combination with distillation of minutiae domain knowledge. 

Inspired by the merging of minutiae domain knowledge into deep CNNs, in this work, we explore the first ever use of a Vision Transformer (ViT~\cite{dosovitskiy2020image}) to learn a fixed length embedding from a fingerprint image. Similar to prior work (which utilized CNN building blocks), we build on top of the vanilla ViT with a strategy for incorporating minutiae domain knowledge into the network's parameters. The use of the multi-headed self attention blocks (MHSA~\cite{vaswani2017attention}) built into ViT for the extraction of fingerprint embeddings is somewhat obvious and intuitive: the MHSA blocks are designed to pay attention to some features of the image more than others and this is exactly what we want in fingerprint recognition, namely we want the ViT to focus strongly on minutiae related features. Our empirical results demonstrate that our hypothesis is correct as our ViT model distilled with minutiae domain knowledge outperforms its non-minutiae distilled counterpart in both authentication and search performance. We also show qualitatively that the ViT model learns to attend to minutiae points in the input fingerprint image via several visualizations. Finally, we show that a fusion of the CNN based embeddings together with the ViT based embeddings leads to a nice boost in recognition performance due to the complementary features both extract.

\begin{figure*}[t!]
 \centering
 \includegraphics[scale=0.75]{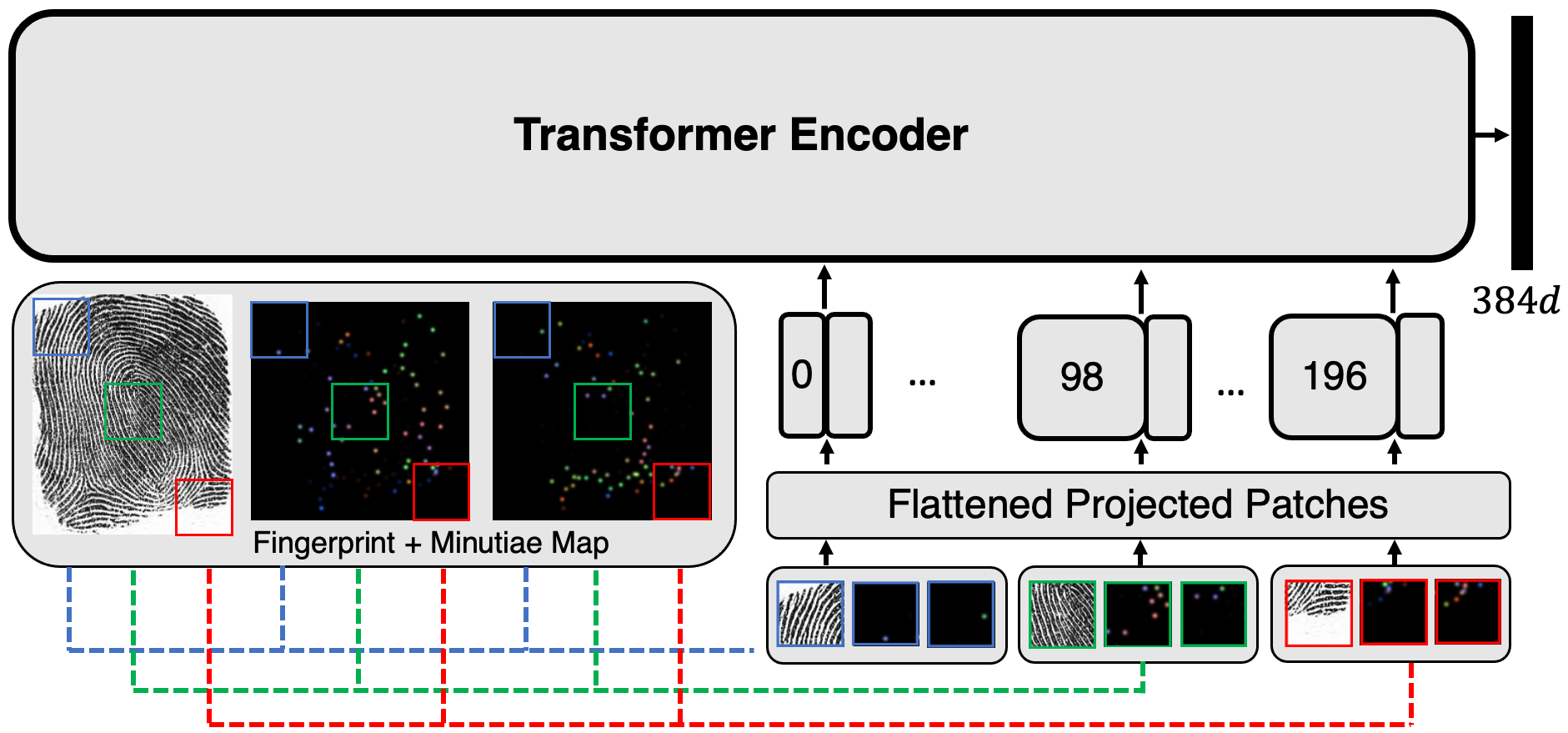}
 \caption{Schematic diagram of the proposed approach. First, a minutiae map for a fingerprint image is extracted (2d heatmap of the minutiae points). Next, patches from the fingerprint image are concatenated with spatially corresponding patches from the 2-channel minutiae map. Finally, these image/minutiae-map concatentated tokens are passed as input to the ViT encoder to extract a fixed-length, 384D fingerprint embedding. Image reproduced from~\cite{dosovitskiy2020image}.}
 \label{fig:schematic}
 \vspace{-1.5em}
\end{figure*}

Concisely, our contributions are as follows:

\begin{itemize}
    \item The first ever use of vision transformers in fingerprint recognition in general and in extracting fingerprint embeddings specifically. We move beyond the vanilla ViT by distilling minutiae domain knowledge into a ViT model\footnote{While this paper was under peer-review, the authors in \cite{tandon2022transformer} posted a manuscript to Arxiv where they used features from a pre-trained CNN model as supervision for learning the weights of a ViT model, rather than letting the ViT model learn it's own representations.}.
    \item A fusion study between ViT based fingerprint embeddings and CNN based fingerprint embeddings to gain more insight into the value of adding a ViT based feature extractor to an existing fingerprint recognition pipeline. 
    \item When fusing minutiae-distilled ViT with a CNN, we approach levels of accuracy of a SOTA commercial matcher on the NIST N2N dataset (TAR=94.23\% @ FAR=0.1\% vs. TAR=96.71\% @ FAR=0.1\%), (Rank-1 search = 97.55\% vs. 99.80\%), while being able to match at orders of magnitude faster speeds due to our fixed length embeddings (2.5M matches/sec vs. 50K matches/sec). 
    \item We will release our training and inference code upon acceptance for further research~\url{github.com/tba}.
\end{itemize}

\section{Related Work}

\subsection{Fingerprint Embeddings}

The research on learning deep fingerprint embeddings has picked up interest, but lags behind the research in learning deep face embeddings. In particular, only a handful of papers have even investigated the feasibility of learning a fixed length embedding for a fingerprint via a deep network~\cite{cao2017fingerprint, song2017fingerprint, song2019aggregating, li2019learning, takahashi2020fingerprint, gu2022latent}.

The existing works on fingerprint embeddings share the common element of using a CNN architecture to learn and extract the embedding. However, there are also a few main differences between the works as well. Some of the initial studies~\cite{cao2017fingerprint, song2017fingerprint} either naively train an off the shelf CNN architecture to extract the embeddings, or they take an additional step in extracting embeddings from multiple scales (to capture the analogous multi-scale features present in fingerprint images of singularities, ridge-flow, minutiae-points, and pores). This approach was also taken up again by~\cite{gu2022latent}, the first paper to explore learning a fingerprint embedding from latent~\footnote{Latent fingerprints or fingermarks are prints left on surfaces, \textit{e.g.}, at a crime scene. They are of very low quality, heavily distorted and lack uniform ridge clarity.} fingerprints. These approaches do not explicitly utilize domain knowledge (\textit{e.g.} minutiae points) to guide the training of the CNN model.

An orthogonal direction that developed to learn a fingerprint embedding was to employ a two-stage process~\cite{song2019aggregating}. In the first stage, minutiae points and their associated local descriptors are extracted from a fingerprint image. In the second stage, these descriptors are aggregated into a single embedding via a 1D CNN aggregation network. A limitation of this approach is that the final representation is limited by the discriminative power of the local descriptors. Furthermore, it requires a relatively computationally expensive two-stage process that must run in series by design. The authors in~\cite{li2019learning} follow a similar vein of thought in learning a global embedding from local patches and then aggregating the patch embeddings via global average pooling.

Finally, in~\cite{engelsma2019learning}, a somewhat hybrid approach of the aforementioned was developed. In particular a single CNN architecture was used to embed a fingerprint image to 192 dimensions. The authors showed that by guiding the network to extract features related to minutiae (via a multi-task network to simultaneously embed the fingerprints and detect minutiae), the fingerprint embeddings could be made more discriminative and generalizable. The authors in~\cite{takahashi2020fingerprint} followed up on this approach via additional tasks in the multi-task learning framework.

In this work, we most closely follow the framework established in~\cite{engelsma2019learning} to use a single network to learn a fingerprint embedding, but to guide that network to extract minutiae related features. In contrast to all the foregoing works which used CNNs to embed the fingerprints, we are the first to propose using a Vision Transformer for the task. We demonstrate how to incorporate the minutiae domain knowledge into the ViT, and we show experimentally that the ViT can extract features which are complementary to the CNN methods such as~\cite{engelsma2019learning}. Our experimental results show the ability of the ViT model, infused with domain knowledge, to surpass the CNN model of~\cite{engelsma2019learning} (also infused with domain knowledge) and we further show that the fusion of the two approaches can lead to even higher performance. 

\subsection{Transformers in Biometrics}

In this study we demonstrate how vision transformers can be guided by minutiae related domain knowledge. We then compare and contrast experimentally the benefits of the transformer models for fingerprint recognition over the baseline CNN methods. Intuitively, we posit that the multi-headed self attention blocks of the transformer models are of great value for learning fingerprint embeddings since fingerprints have some local features (minutiae points) which should be attended to with much more weight than other more global features (e.g., ridge flow or singularity points). 

We do note that although transformers have not been utilized as of yet for fingerprint recognition, they have been introduced with some success as an approach for face recognition and expression analysis~\cite{zhong2021face, jacob2021facial, xue2021transfer, zhang2022transformer, zheng2022general}

\section{Approach}

Our approach consists of i) a strategy for incorporating fingerprint domain knowledge (minutiae) into the ViT architecture, ii) training the ViT on a combination of synthetic and real fingerprints, and iii) fusing the trained ViT with a SOTA CNN based fingerprint embedding model. These steps are explained further below.

\subsection{Minutiae Guided Vision Transformers}

The vanilla ViT proposed in~\cite{dosovitskiy2020image} takes as input a 2D image of size $h\times~w$ and and raster scans a set of $N$ patches (also known as tokens denoted as $x_i$) where each token is of size $n~\times~n$. In our case, we preprocess the fingerprints to be of size $224~\times~224$ and we set the patch size to $16~\times~16$, or $256d$ after flattening. Therefore, the number of tokens input to our ViT model is $N=196$, each of dimension $256$. Ordinarily, these flattened image patches would then be linearly projected and passed to the MHSA blocks~\cite{vaswani2017attention} of the transformer encoder. However, in our case, rather than directly passing raw, flattened image pixels to the ViT encoder, we additionally concatenate the minutiae points located within each patch to the respective input token. By adding this additional input signal, our aim is to guide the ViT to attend more specifically to minutiae points present in the raw pixels rather than other less discriminative features.

Inserting minutiae points into the ViT architecture is not trivial since the number of input tokens is fixed (e.g., $196$), but the number of minutiae points is variable and unordered. In other words, we cannot simply concatenate the $(x_i, y_i, \theta_i)$ tuple for each minutiae point to our tokens $x_i\in\mathbb{R}^{256}$ because we would end up with tokens of varying dimension depending on how many minutiae points were present in a given patch (some patches may have 0 minutiae, while others can have multiple minutiae). To solve this problem, we relay on the minutiae-map first proposed in~\cite{cao2019end} and also used in~\cite{engelsma2019learning}. The minutiae map takes an input an unordered, variable length, minutiae set $T=\{(x_1, y_1, \theta_1), ..., (x_m, y_m, \theta_m)\}$ with $m$ minutiae points, extracted from a fingerprint of size $h, w$ and converts it into a 3D heatmap $H$ of size $h\times~w~\times~c$. In this heatmap, the hot-spots at a position $(i, j, c)$ indicate the presence of a minutiae point located at $(i, j)$ with an orientation determined by the channel $c$. If we set $c=1$, then no orientation information is encoded. As we add more channels, more granular orientation values can be recovered. To balance computational expense with orientation granularity, we set $c=2$ in our experiments.

\begin{comment}
\begin{figure}[t!]
 \centering
 \includegraphics[scale=0.3]{figures/mmap.png}
 \caption{A minutiae map~\cite{cao2019end} is a sparse, fixed-size, 3D representation of a variable length minutiae set (key-point set). The hot-spots indicate the location of a minutiae-point. The orientation of the minutiae point is encoded by the channel.}.
 \label{fig:mmap}
\end{figure}
\end{comment}

\begin{figure}[t!]
 \centering
 \includegraphics[width=\linewidth]{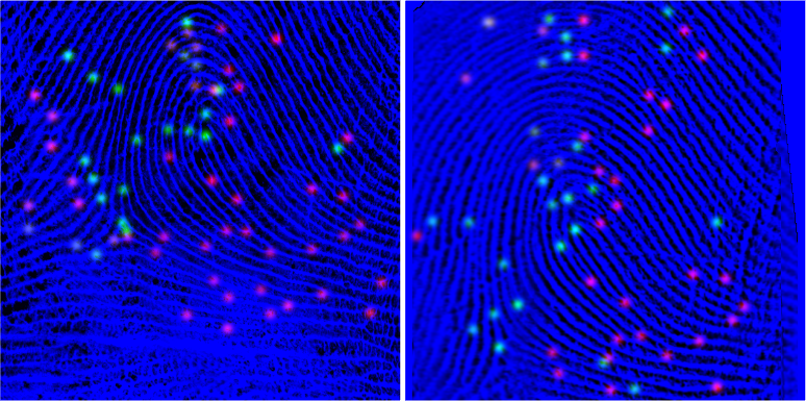}
 \caption{Example minutiae and fingerprint image concatenations represented as RGB images, where the blue channel contains the grayscale fingerprint ridge structure, the green channel contains minutiae locations which have orientation with the range [0, 180) and the red channel contains minutiae locations which have orientation with the range [180, 360).}.
 \label{fig:ex_concats}
 \vspace{-1.5em}
\end{figure}

\begin{comment}
\begin{figure*}[!t]
\begin{minipage}{.2\textwidth}
\centering
\includegraphics[scale=.2]{figures/printsgan/0-1.jpg}
\end{minipage}%
\begin{minipage}{.2\textwidth}
\centering
\includegraphics[scale=.2]{figures/printsgan/1-1.jpg}
\end{minipage}
\begin{minipage}{.2\textwidth}
\centering
\includegraphics[scale=.2]{figures/printsgan/2-1.jpg}
\end{minipage}%
\begin{minipage}{.2\textwidth}
\centering
\includegraphics[scale=.2]{figures/printsgan/3-1.jpg}
\end{minipage}%
\begin{minipage}{.2\textwidth}
\centering
\includegraphics[scale=.2]{figures/printsgan/4-1.jpg}
\end{minipage}

\par\medskip

\begin{minipage}{.2\textwidth}
\centering
\includegraphics[scale=.2]{figures/printsgan/0-2.jpg}
\end{minipage}%
\begin{minipage}{.2\textwidth}
\centering
\includegraphics[scale=.2]{figures/printsgan/1-2.jpg}
\end{minipage}%\par\medskip
\begin{minipage}{.2\textwidth}
\centering
\includegraphics[scale=.2]{figures/printsgan/2-2.jpg}
\end{minipage}%
\begin{minipage}{.2\textwidth}
\centering
\includegraphics[scale=.2]{figures/printsgan/3-2.jpg}
\end{minipage}%
\begin{minipage}{.2\textwidth}
\centering
\includegraphics[scale=.2]{figures/printsgan/4-2.jpg}
\end{minipage}%
\centering
\caption{Example rolled fingerprint impressions from the PrintsGAN synthetic fingerprint database~\cite{engelsma2022printsgan}. Each column is a different identity (finger) while each row shows a different impression of the same identity}
\label{fig:example_images}
\end{figure*}
\end{comment}

With a minutiae map $H$ extracted from a given fingerprint, we can now concatenate minutiae points to each input patch/token prior to extracting an embedding with ViT (schematic Figure~\ref{fig:schematic}) (full image concatenated with the two-channel minutiae map shown in Figure~\ref{fig:ex_concats}). In particular, for a $16~\times~16$ token extracted from the fingerprint image at position $(i, j)$, we can extract the corresponding minutiae points from $H$ by extracting two (since $c=2$) $16\times~16$ patches at positions $(i, j)$ from $H$. These three $16~\times~16$ patches (one from the image, and the other from the two minutiae-map channels), can then be concatenated and flattened into a single token of size $256\times3$ and fed as input to the ViT architecture. Similar to~\cite{dosovitskiy2020image}, a class embedding is used to learn a representation of the image (the class embedding is fed into a softmax cross-entropy loss function). In our case, the ViT's class embedding needs to be able to classify the input fingerprint image into it's identity (each finger in the dataset is treated as a unique identity or class). After training, the classification layer can be discarded, the previous embedding layer of dimension $384d$ can be used as our fixed-length representation.

\subsection{Datasets}

A primary reason for the dearth of research on the topic of fixed-length fingerprint representations via deep networks is due to limited large scale fingerprint data needed to train deep fingerprint embedding models. However, advancements in generative models have now led to the public release of synthetic fingerprints~\cite{engelsma2022printsgan} which have been demonstrated to facilitate learning fixed length embeddings when combined with real fingerprint data such as that released publicly by NIST in the SD 302 dataset~\cite{sd302}. Therefore, in our experiments, we follow the approach of~\cite{engelsma2022printsgan} in first pretraining our models with synthetic fingerprints and subsequently fine-tuning them with a portion of the NIST SD302 real fingerprints. Note, when fine-tuning the pretrained models, we fine-tune all parameters with the same optimizer, learning rate, and scheduler used to pretrain the networks (the synthetic data is used only to provide a better initialization to the model weights in a similar manner to how ImageNet~\cite{deng2009imagenet} data is often used to pretrain image classification models.) 

The synthetic data used in our experiments (dubbed PrintsGAN data) comes from a set of $525K$ synthetic fingerprints publicly released~\footnote{\url{https://biometrics.cse.msu.edu/Publications/Databases/MSU_PrintsGAN/}} by the authors of~\cite{engelsma2022printsgan}. The dataset statistics are summarized in Table~\ref{table:printsgan} along with the training, validation, and testing splits which we created for the data.

The real fingerprint data used in our experiments comes from the publicly available NIST SD-302 dataset. Of the popular NIST datasets, both NIST SD4 and NIST SD14 have been rescinded, leaving NIST SD-302 as the only remaining NIST dataset publicly available for our use-case. While other datasets for fingerprint do exist in the public domain (e.g., FVC datasets), i) they are very small scale (approximately 1K fingerprints) and ii) our Internal Review Board (IRB) does not support their use. To account for the fact that our evaluations are on a single real dataset, we perform 5-fold cross validation for both our authentication and search experiments, reporting both accuracy and also standard deviation across the different folds. The dataset statistics of the SD-302 dataset, often referred to as the Nail-to-Nail or N2N dataset, are listed in Table~\ref{table:sd302}. We note that the N2N dataset is very challenging because the fingerprints were captured using 18 different fingerprint readers (some of them experimental prototypes which capture lower quality fingerprints). Example synthetic fingerprints from PrintsGAN are shown in our qualitative experimental results in Figure~\ref{fig:occluded}. We do not show any example images from the N2N real fingerprint dataset due to privacy concerns. 

\begin{table}[t]
\caption{PrintsGAN Synthetic Fingerprint Dataset\tnote{1}}
 \centering
\begin{threeparttable}
\begin{tabular}{c c c}
 \toprule
 \specialcell{Number of \\Fingers \\(Identities)} & \specialcell{Avg. Impressions \\per Finger} & \specialcell{Number of \\Images} \\
 \toprule
%  \multicolumn{3}{c}{Full Dataset} \\
 34,985 & Full Dataset (15) & 525K \\
\midrule
%  \multicolumn{3}{c}{Train Partition} \\
 28,344 & Train Partition (15) & 425K  \\
 \midrule
%  \multicolumn{3}{c}{Validation \& Test Partitions} \\
 3,142 & Validation Partition (15) & 47K  \\
 \midrule
 3,499 & Test Partition (15) & 52K  \\
 \bottomrule
\end{tabular}
\begin{tablenotes}
\item[1] PrintsGAN produces rolled synthetic fingerprints.
\end{tablenotes}
\end{threeparttable}
\label{table:printsgan}
\end{table}

\begin{table}[t]
\caption{NIST SD-302 (N2N) Real Fingerprint\tnote{1} Dataset}
 \centering
\begin{threeparttable}
\begin{tabular}{c c c}
 \toprule
 \specialcell{Number of \\Fingers \\(Identities)} & \specialcell{Avg. Impressions \\per Finger} & \specialcell{Number of \\Images} \\
 \toprule
%  \multicolumn{3}{c}{Full Dataset} \\
 2,000 & Full Dataset (12) & 25K \\
\midrule
%  \multicolumn{3}{c}{Cross Validation Train Fold} \\
 1600 & \specialcell{Cross Validation \\Train Fold (12)} & 20K  \\
 \midrule
%  \multicolumn{3}{c}{Cross Validation Val/Test Fold} \\
 200 & \specialcell{Cross Validation \\Val \& Test Folds (12)} & 2.5K  \\
 \bottomrule
\end{tabular}
\begin{tablenotes}
\item[1] The NIST SD-302 partition we use is comprised of rolled and plain-print impressions.
\end{tablenotes}
\end{threeparttable}
\label{table:sd302}
\end{table}

\section{Experimental Results}

Our baseline methods which we compared against our minutiae-distilled ViT are comprised of i) a ResNet-50 CNN model utilizing the additive angular margin loss~\cite{deng2019arcface}, ii) DeepPrint, a SOTA CNN approach proposed by~\cite{engelsma2019learning} for fingerprint embeddings, and iii) a SOTA fingerprint SDK Verifinger v12.3, which includes both a minutiae-based matcher and a proprietary matching algorithm that is undisclosed. DeepPrint and Verifinger results are taken from~\cite{cross-domain-fp}. We use the same train, validation, and test splits as~\cite{cross-domain-fp} to enable fair comparison of our method with these results.

\subsection{Authentication Experiments}

First, we report authentication (1:1 matching) results (Table~\ref{table:verification}). Genuine scores are computed via pairwise comparisons between all of the impressions of the same finger and imposter scores are computed by matching each fingerprint image of one finger to all other impressions of every other finger. For reference, $15,143$ genuine and $3,229,735$ imposter matches were computed for the test set of the first split of NIST SD 302. 

From the authentication results, we first note that by adding a minutiae prior into the ViT model (ViT Concat), via concatenated inputs of the grayscale fingerprint ridge structure and 2-channel minutiae maps, the TAR is increased over the vanilla ViT from $90.11\%$ to $90.96\%$ and the standard deviation in TAR across test splits reduces from $3.26\%$ to $2.23\%$.

Next, we note from Table~\ref{table:verification} that by fusing (score level fusion) the transformer based ViT models with CNN based models DeepPrint and ResNet, appreciable performance gains can be observed. This stands in contrast to when two CNN models are fused together. A major motivation of this study is to show that the disparate architecture and building blocks of the ViT could better complement an existing CNN architecture via fusion than a second CNN. 

Finally, we acknowledge that our best performing configuration of ResNet-50 fused with ViT-Concat is not yet at parity with Verifinger (TAR of 94.23\% vs. TAR of 96.71\%), one of the best performing fingerprint matchers. However, our results make tremendous strides in closing the gap between the SOTA published method DeepPrint (TAR of 82.11\%) and Verifinger (reducing the error rate between DeepPrint and Verifinger by 83\%). Furthermore, we posit that with a larger training dataset, as is the case with the DeepPrint model pretrained on MSP vs. DeepPrint pretrained on PrintsGAN (and most likely the case with Verifinger), our performance would also improve further.

\begin{table*}[t]
\caption{Authentication (1:1 matching) performance on five disjoint test splits of NIST SD 302.}
\centering
\begin{threeparttable}
\begin{tabular}{c c c c c c}
\toprule
Model & \begin{tabular}[c]{@{}c@{}}Number of \\Train Images\end{tabular} & \begin{tabular}[c]{@{}c@{}}Number of\\Parameters\end{tabular} & \begin{tabular}[c]{@{}c@{}}Inference Speed (Nvidia Tesla\\V100-SXM2=16GB.)\end{tabular} & TAR ($\%$) @ 0.1$\%$ FAR \\
\toprule
\begin{tabular}[c]{@{}c@{}}Verifinger proprietary \\matcher\end{tabular} & Unknown & N/A & \textbf{600ms}\tnote{1} & $\mathbf{96.71\pm1.16}$ \\
\midrule
\begin{tabular}[c]{@{}c@{}}Verifinger ISO \\matcher\end{tabular} & Unknown & N/A & 600ms\tnote{1} & $95.09\pm1.59$ \\
\toprule 
\begin{tabular}[c]{@{}c@{}}ResNet50, (ArcFace Loss)\end{tabular} & \begin{tabular}[c]{@{}c@{}}445K\tnote{4}\end{tabular} & 25.56M & 10.5ms & $93.54\pm1.66$ \\
\midrule
\begin{tabular}[c]{@{}c@{}}DeepPrint-A, (Softmax CE loss)\end{tabular} & \begin{tabular}[c]{@{}c@{}}873K\tnote{5}\end{tabular} & 76.93M & 40.4ms & $90.60\pm1.72$ \\
\midrule
\begin{tabular}[c]{@{}c@{}}DeepPrint-B, (Softmax CE loss)\end{tabular} & \begin{tabular}[c]{@{}c@{}}445K\tnote{4}\end{tabular} & 76.93M & 40.4ms & $82.11\pm2.96$ \\
\midrule
ViT & \begin{tabular}[c]{@{}c@{}}445K\tnote{4}\end{tabular} & 22.0M & 10.2ms & $90.11\pm3.26$ \\
\midrule
ViT Concat\tnote{2} & \begin{tabular}[c]{@{}c@{}}445K\tnote{4}\end{tabular} & 22.0M & 25.2ms & $90.96\pm2.23$ \\
\toprule
\begin{tabular}[c]{@{}c@{}}ResNet50 + DeepPrint-B\tnote{3}\end{tabular} & \begin{tabular}[c]{@{}c@{}}445K\tnote{4}\end{tabular} & 102.5M & 50.9ms & $93.56\pm1.63$ \\
\midrule
\begin{tabular}[c]{@{}c@{}}ResNet50 + ViT\tnote{3}\end{tabular} & \begin{tabular}[c]{@{}c@{}}445K\tnote{4}\end{tabular} & 47.56M & 20.7ms & $94.20\pm1.61$ \\
\midrule
\begin{tabular}[c]{@{}c@{}}ResNet50 + ViT Concat\tnote{2,3}\end{tabular} & \begin{tabular}[c]{@{}c@{}}445K\tnote{4}\end{tabular} & 47.56M & \textbf{35.7ms} & $\mathbf{94.23\pm1.54}$ \\
\midrule
\begin{tabular}[c]{@{}c@{}}ViT + DeepPrint-B\tnote{3}\end{tabular} & \begin{tabular}[c]{@{}c@{}}445K\tnote{4}\end{tabular} & 98.93M & 50.6ms & $90.45\pm3.00$ \\
\midrule
\begin{tabular}[c]{@{}c@{}}ViT Concat\tnote{2}  + DeepPrint-B\tnote{3}\end{tabular} & \begin{tabular}[c]{@{}c@{}}445K\tnote{4}\end{tabular} & 98.93M & 65.6ms & $90.94\pm2.25$ \\
\midrule
\begin{tabular}[c]{@{}c@{}}ViT Concat\tnote{2}  + ViT\tnote{3}\end{tabular} & \begin{tabular}[c]{@{}c@{}}445K\tnote{4}\end{tabular} & 44.0M & 35.4ms & $91.35\pm2.19$ \\
\bottomrule
\end{tabular}
\begin{tablenotes}
\item[1] 600ms for template extraction on at least an Intel Core 7-8xxx family processor.
\item[2] Uses grayscale fingerprint images concatenated with 2 channel minutiae maps.
\item[3] ($w_1=0.7$, $w_2=0.3$) where $w_1$ and $w_2$ are the fusion weights of the first and second model, respectively.
\item[4] Training data consists of PrintsGAN~\cite{engelsma2022printsgan} + NIST SD 302~\cite{sd302} datasets.
\item[5] Training data consists of MSP~\cite{yoon2015longitudinal} + NIST SD 302~\cite{sd302} datasets.
\end{tablenotes}
\end{threeparttable}
\label{table:verification}
\end{table*}

\subsection{Search Experiments}
We report both the closed-set (each probe has a mate in the gallery) and open-set (some probes do not have a mate in the gallery) search. Closed-set is reported via the CMC curve (\% of times true mate is in the top-N for all probes) and open-set is reported via a plot of False Positive Identification Rate (FPIR) vs. False Negative Identification Rate (FNIR).  

\subsubsection{Closed-Set Search}
For closed-set search, the gallery construction is comprised of two impressions from each finger in the PrintsGAN test partition, yielding $6,998$ fingerprint images. In addition, we randomly selected two impressions from each finger in the NIST SD302 test partition to serve as enrollments and set aside the other impressions as probes. In total, the gallery consisted of $7,398$ total fingerprint images, whereas the number of probe images varies per cross-validation split but is roughly around $2,000$ probe fingerprint images ($200$ fingers $\times 10$ impressions). For reference, the probe set for the first cross validation split consisted of $2,148$ probe fingerprint images.

\begin{figure}[t!]
 \centering
 \includegraphics[width=0.9\linewidth]{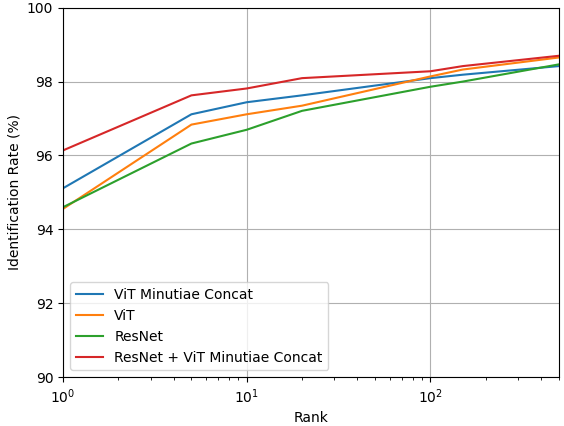}
 \caption{Cumulative Match Characteristic (CMC) curve for various fingerprint recognition models (ResNet and ViT, and the fusion of the two) on a gallery size of $7,398$ fingerprint images. (Higher curve is better.)}
 \label{fig:cmc}
 \vspace{-1.5em}
\end{figure}

The closed-set search results are shown in Figure~\ref{fig:cmc}. From these results, we note that the minutiae distilled ViT again outperforms the vanilla ViT. Furthermore, it outperforms the best CNN based baseline (ResNet-50). Finally, we again demonstrate that by fusing ViT with the CNN, we obtain even better results than the two stand-alone. 

\begin{figure}[t!]
 \centering
 \includegraphics[width=\linewidth]{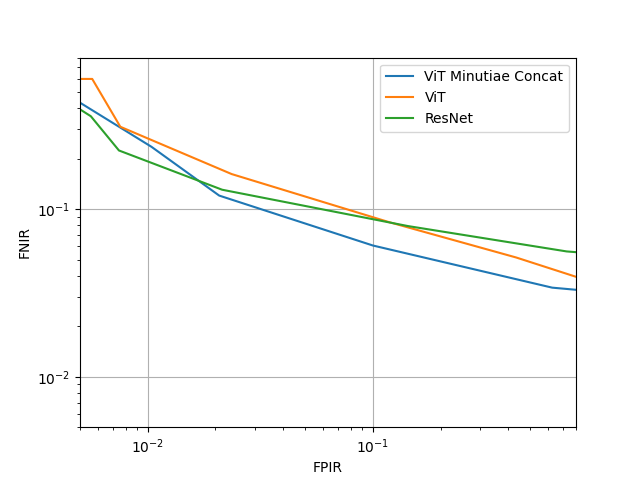}
 \caption{FNIR vs. FPIR for various fingerprint recognition models. (Lower curve is better.)}
 \label{fig:det}
 %\vspace{-1.5em}
\end{figure}

\subsubsection{Open-Set Search}
To construct the gallery for the open-set scenario, we start off with the same gallery as was used in the closed-set scenario. Then, we remove the true mates from the gallery for half of the probe subjects which were used in the closed-set setting. This results in $100$ mated probes (probes which have a genuine match present in the gallery) and $100$ un-mated probes (probe subjects which do not have a corresponding genuine match in the gallery). For each split, the size of the gallery is therefore reduced by $200$ fingerprint images ($2$ impressions per $100$ probe fingers removed from the gallery), resulting in a gallery size of $7,198$. 

The open-set identification performance is shown in Figure~\ref{fig:det}. We observe that in this case, the ResNet outperforms the ViT models at strict operating points, however, the minutiae-distilled ViT outperforms the vanilla ViT and bests the ResNet at many FPIR operating points. 

% \subsection{Decorrelation Study}

% show the prediction decorrelation experiment we discussed - how many times does the 2nd candidate agree between two models? the less times the better. explain how this is computed.

\subsection{Qualitative Analysis}
\begin{figure}[t!]
 \centering
 \includegraphics[width=\linewidth]{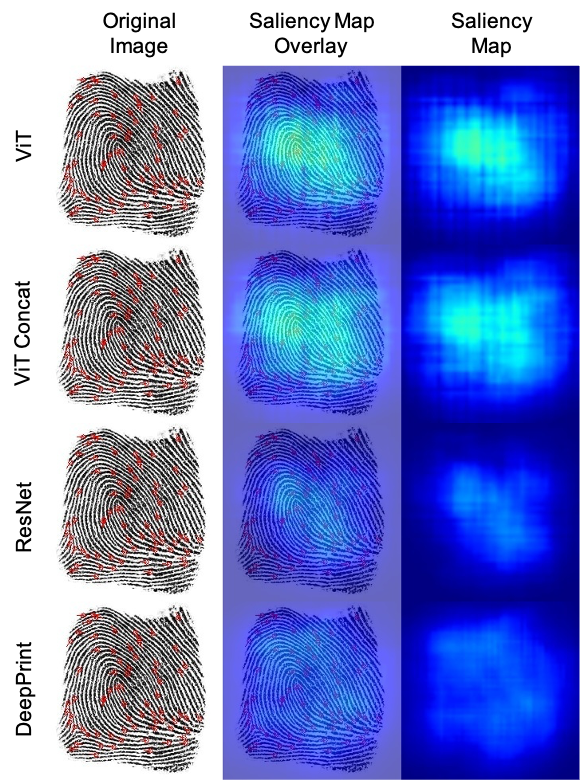}
 \caption{Saliency map visualization of ViT, ViT Concat (ViT on concatenated fingerprint ridge structure and 2-channel minutiae map inputs), ResNet, and DeepPrint.}
 \label{fig:saliency}
 \vspace{-1.5em}
\end{figure}

To better understand the nature of the features extracted by the ViT and the CNN models, we visualized the saliency maps for both (Figure~\ref{fig:saliency}). The ViT saliency map (top row) compared to ResNet (third row) shows that the ViT model is using more area of the image to make its prediction, whereas the ResNet CNN model is ignoring a large proportion of the image. Furthermore, comparing ViT to ViT Concat and ResNet to DeepPrint, it appears that providing minutiae supervision during training (in the case of ViT Concat and DeepPrint) also forces the model to focus on more of the fingerprint area. The fact that the saliency maps for ViT and ResNet are more different compared to ResNet and DeepPrint, also suggests that ViT and CNN models are focusing on complementary information, one focusing on larger, more central features and one looking at more local regions throughout the image. Thus, the fusion of the two model architectures leads to a performance increase. 

\begin{figure}[t!]
  \centering
  \subfloat[Failure Case 1]{\includegraphics[width=0.9\linewidth]{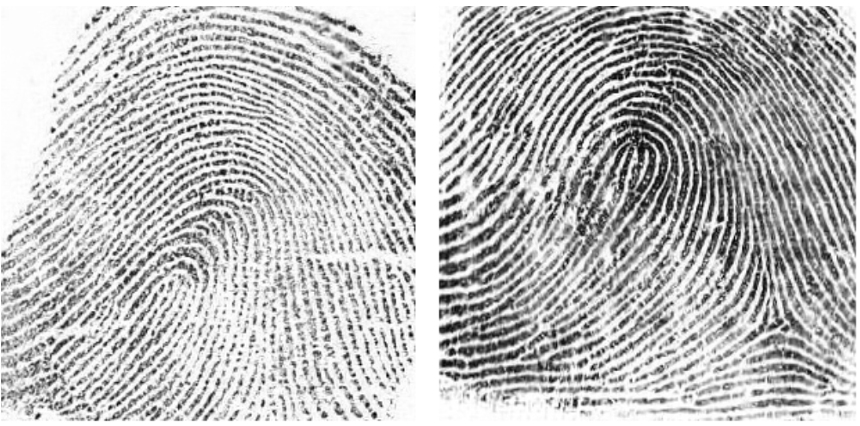}\label{fig:f1}} \\
  \hfill
  \subfloat[Failure Case 2]{\includegraphics[width=0.9\linewidth]{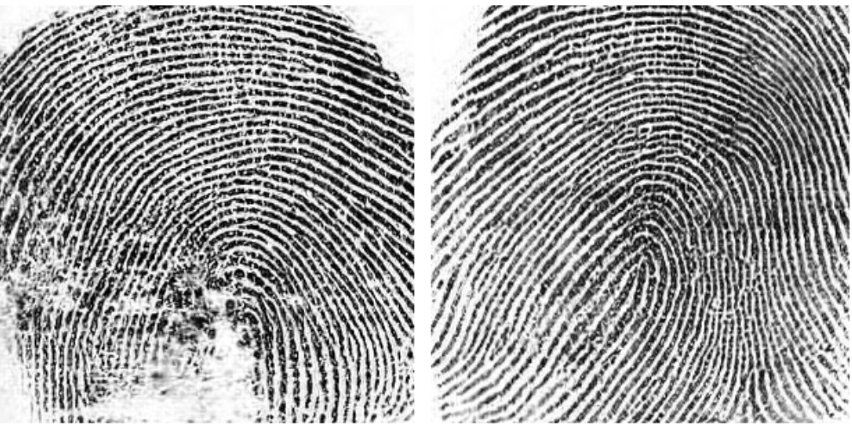}\label{fig:f2}}
  \caption{vanilla-ViT failed, minutiae-distilled ViT successfully matched.}
  \label{fig:occluded}
\end{figure}

Finally, to better understand the benefit of incorporating minutiae domain knowledge into ViT, we have displayed two failure case examples where the two genuine fingerprint images are not matched by ViT but are correctly matched by ViT with concatenated inputs. See Figure~\ref{fig:occluded}. These failure cases suggest that in case of large occlusions, providing the minutiae supervision helps to guide the network to focus on the local features which are not entirely occluded, thus enabling a true match.

\section{Matching Time}

The major advantage of our deep fingerprint embeddings over the commercial minutiae (key-point) matcher comes at search time. Our embeddings enable comparing 2.5 million fingerprints per second vs. the commercial system of 50K matches/second (computed on an Intel i7 processor). 

\section{Conclusion}
In this paper, we explored the use of ViT models for fingerprint recognition and showed that the standalone performance of ViT for fingerprint authentication and identification is comparable to CNN-based models, and in some cases, exceeds the performance of CNN models, such as  ResNet50 and DeepPrint, while using fewer number of parameters. Furthermore, we showed that embedding minutiae domain knowledge into the input of ViT leads to an increase in performance, which is analogous to previous research incorporating minutiae domain knowledge into CNNs (e.g., DeepPrint). This motivates further investigation into incorporating fingerprint domain knowledge into ViT to improve the recognition performance even further. Lastly, we showed the the features learned by ViT are complementary to the features learned by CNN methods, thus the fusion of the two models obtains significant performance increases, approaching that of state-of-the-art fingerprint recognition systems such as Verifinger v12.3, with significantly faster matching.

{\small
\bibliographystyle{ieeetr}
\bibliography{egbib}
}

\end{document}